\documentclass{svproc}
\usepackage{url}
\usepackage{multicol}
\usepackage{graphicx}
\usepackage{subcaption}
\usepackage{multirow}
\usepackage{float}
\usepackage[bookmarks=true]{hyperref}
\usepackage{amsmath}
\usepackage{capt-of}
\usepackage{color}
\usepackage{todonotes}
\usepackage{url}
\usepackage{ulem}
\usepackage{balance}

\begin{document}
\mainmatter             
\title{Sensorless Pose Determination using Randomized Action Sequences}
\titlerunning{Sensorless Pose Determination using Randomized Action Sequences}  
%
\author{Pragna Mannam\inst{1} \and Alexander Volkov Jr.\inst{1}
Robert Paolini\inst{1} \and \\Gregory Chirikjian\inst{2} \and Matthew T. Mason\inst{1}}
\authorrunning{Pragna Mannam et al.} 
\tocauthor{Pragna Mannam, Alexander Volkov Jr., Robert Paolini, Gregory Chirikjian, and Matthew T. Mason}

\institute{The Robotics Institute, Carnegie Mellon University, Pittsburgh,  PA 15213, USA
\and
Department Of Mechanical Engineering, Johns Hopkins University,\\
Baltimore, MD 21218, USA \\
\email{pmannam@andrew.cmu.edu, avolkovjr@cmu.edu, rpaolini@cmu.edu, gchirik1@jhu.edu, matt.mason@cs.cmu.edu}}

\maketitle              

\begin{abstract}
This paper is a study of 2D manipulation without sensing and planning, by exploring the effects of unplanned randomized action sequences on 2D object pose uncertainty. Our approach follows the work of Erdmann and Mason's sensorless reorienting of an object into a completely determined pose, regardless of its initial pose. While Erdmann and Mason proposed a method using Newtonian mechanics, this paper shows that under some circumstances, a long enough sequence of random actions will also converge toward a determined final pose of the object. This is verified through several simulation and real robot experiments where randomized action sequences are shown to reduce entropy of the object pose distribution. The effects of varying object shapes, action sequences, and surface friction are also explored. 
\end{abstract}

\section{Introduction} 
\label{intro}
Robots are envisioned to manipulate and interact with objects in unscripted environments and accomplish a diverse set of tasks.
Towards this, reducing object pose uncertainty is necessary for successful task execution.
There are natural ways to reduce pose uncertainty including the addition of physical constraints, relative positioning to a known object's pose, and actively sensing the desired object's pose.
In this paper, we explore a novel pose uncertainty reduction technique based on executing randomized sequence of actions.
We evaluate our proposed pose uncertainty reduction technique on parts orienting, an industrial automation task.

Reducing task state uncertainty in parts orienting systems is an important part of factory automation, especially product assembly.   
The problem is to take parts in a disorganized jumble and to present them one at a time in a predictable pose. 
Most industrial solutions involve a part-specific mechanical design.   
One goal of parts-orienting research is to avoid part-specific mechanical designs, reducing the time required to develop the automation for a new or redesigned product.   

\begin{figure}[t!]
\centering
\includegraphics[width=0.55\linewidth]{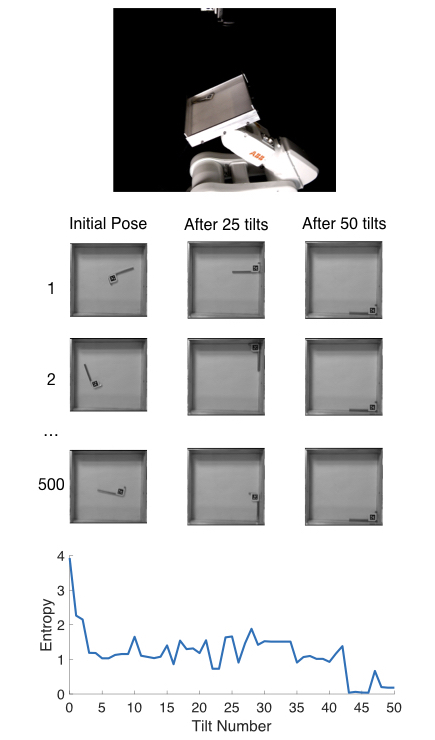}
\caption{Experimental setup (top).  An industrial robot tilts an allen key, with April Tag attached, in an aluminum tray. The overhead camera records the pose of the allen key after each tilt. Each trial (1), (2), $\dots$(500) performs the same random sequence of actions with a different initial position. The pose before the sequence, mid-sequence after 25 tilts, and after the sequence of 50 tilts are shown per trial, as well as the entropy of object pose distribution over 500 total repeated trials (bottom). }
\label{fig:intro}
\end{figure}

Tray-tilting is one kind of part-agnostic object reorientation system. The original tray-tilting work was an early entry in a research approach termed ``minimalism.''  
Minimalism refers to ``the art of doing X without
Y,'' or ``finding the minimal configuration of resources to solve a task" \cite{Bohringer1997}.
The purpose of the approach is not just to conserve resources, but to yield insights into the structure of tasks and the nature of perception, planning, and action.

The role of sensing in the sense-plan-act structure was examined by the tray-tilting work of Erdmann and Mason~\cite{M_E}, which eliminated all uncertainty in the task state without sensing.  
Rather than sensing, the discrete set of feasible task states could sometimes be reduced to a singleton through judicious choice of actions.  
So in some tasks, even allowing for the noisy mechanics of frictional contact, task state uncertainty can be eliminated without sensing.

While the original paper by Erdmann and Mason~\cite{M_E} examined the role of sensing, this paper is an extension exploring similar minimalism in planning.
We replace Erdmann and Mason's ~\cite{M_E} planned sequence of actions with a randomized sequence of actions and evaluate the reduction in object pose uncertainty.
Using tray-tilting random actions, instead of planning, can provide simple part-agnostic designs in factory automation.
Our experiments stay close to the original work to focus on the role of planning. 
We test the limits of minimalism with respect to system complexity and hope to pursue its practical applications in future work.
For this reason, we adopt the same task domain:  planar sliding of a laminar object in a rectangular tray.
The robot can tilt the tray as desired, and the goal is to move the object to a single final pose, irrespective of its initial pose. 
If independent actions do not scramble the task state too much, then occasionally some action maps two initial task states to the same final task state.
Furthermore, we expect the set of feasible task states to approach a singleton, for sufficiently long sequences, as seen in Figure \ref{fig:intro}.
The phenomenon, while also reminiscent of contraction mapping, is similar to an interesting card trick called the Kruskal Count~\cite{kruskal}, so we have dubbed the phenomenon as ``Kruskal effect." 

The goal of this paper is to better understand the role of planning by observing the effects of using only randomized actions. 
For proof of concept, we experimented with various triangular objects. 
We note that orienting a symmetrical or concave shape with this approach might be more difficult.
For objects similar to allen keys, relatively low tray friction noise, and a long enough sequence of random actions, we show that the Kruskal effect applies. We also observe that it does not apply as well to cases with high tray friction noise, and exploration into more cases is left for future work.
The insights we gain from our exploration of the limits of Kruskal effect can lay the foundations for compartmentalized tray-tilting of a kit of parts in factory automation or 3D pose determination in future work. 

\subsection{Previous work}
The problem of presenting a single object from disorganization has interested robotics researchers as far back as Grossman and Blasgen's work in 1975 \cite{Grossman1975}.  
Grossman and Blasgen introduced a fixed tilted tray that used vibration to eliminate the effects of friction.  
An irregular part in the tray would settle into one of a small number of stable poses, and the robot used a touch probe to disambiguate the pose.
V{\'a}rkonyi \cite{Varkonyi2014} includes additional details on approaches to the problem by using simulation to systematically evaluate various pose estimators. 

Erdmann and Mason~\cite{M_E} substituted a fixed tray with an active tilting tray, and showed that for some parts, a sequence of tilts would reduce the possible poses to a singleton, completely orienting the part without a touch probe or any other sensor.

While the tray is not part-specific, the Erdmann and Mason~\cite{M_E} approach uses part-specific motions.  
In this paper, we substitute the motions with a random sequence of tilts, which is not part-specific.  
If we can identify an interesting class of parts that are oriented by a random sequence, then we have what is sometimes termed a ``universal" parts orienting system.  B\"{o}hringer et al. 
\cite{Bohringer1997} includes an overview of universal parts orienting research, detailing the design and implementation of planar force vector fields that will orient asymmetric laminar parts.
  
Sanderson \cite{SandersonEntropy} introduced parts entropy in the context of automated manufacturing.  
We use probability density functions in the configuration space, $SE(2)$ for planar motion of rigid parts, and we use entropy to measure and compare distributions.
Our calculation of entropy is based on Chirikjian's work on computing the discrete entropy of histograms \cite{Chirikjian2009}.

Pose uncertainty has previously been addressed with the use of action, rather than sensing, in manipulation. 
Brost \cite{brost} uses squeeze-grasp actions to intrinsically reduce uncertainty of the object's position.
Goldberg \cite{Goldberg1993b} planned sequences of pushes and squeezes to orient planar polygons up to symmetry.  
Zhou et al. \cite{Zhou2017} plans similar sequences based on an efficient simulation of planar pushing. 
Berretty et al. \cite{Goldberg} proposed an approach of executing pulling actions using overhead fingers for object reorientation. 
Akella and Mason \cite{AkellaNests} applied a similar approach to parts with uncertain shape.
Unlike these previous works, we use a random sequence of actions to reduce the uncertainty associated with the pose of an object.
A random sequence of actions is a part-agnostic plan that minimizes software complexity and hardware changes for new parts.

\subsection{Paper Overview}
First, Section \ref{sec:entropy} will discuss how we calculate pose uncertainty after every action in the sequences. Then, experimental setup and results are presented in Sections \ref{sec:sim} and \ref{sec:rob}, respectively. Finally, we discuss our observations in Section \ref{sec:disc} and conclude with directions for future research in Sections \ref{sec:conclusion} and \ref{sec:future}. 

\section{Measuring Order: Entropy}
\label{sec:entropy}
Parts entropy describes the probability distribution of an object's pose over repeated tasks ~\cite{SandersonEntropy}.
We measure object pose uncertainty using parts entropy throughout our randomized action sequences over many trials.  
Using parts entropy from Sanderson ~\cite{SandersonEntropy} and notation from Lee et al.~\cite{greg_ent}, we define an object's pose in a tray of size $a \times b$ with the tuple $(x, y, \theta)$ where each coordinate is discretized with uniform spacing such that 
\begin{align}
 x \in \{ x_j : j = 1, \ldots, \alpha \} \phantom{x} \text{on} \phantom{x} \left[ 0, a \right] \label{eq1} \\
 y \in \{ y_k  :  k = 1, \ldots, \beta \} \phantom{x} \text{on} \phantom{x} \left[ 0, b \right] \label{eq2} \\
\theta \in \{ \theta_m  : m = 1, \ldots, \gamma \}  \phantom{x} \text{on} \phantom{x} \left[ 0, 2\pi \right]  \label{eq3} 
\end{align}
The number of discretized intervals are 
\begin{align}
\alpha = \displaystyle \frac{a}{\epsilon_p}, \phantom{xx} \beta = \frac{b}{\epsilon_p}, \phantom{xx} \gamma = \frac{2\pi}{\epsilon_r}   \label{eq4}
\end{align}
where $\epsilon_p$ and $\epsilon_r$ are the positional and rotational resolutions, respectively.
We selected resolutions $\epsilon_p$ and $\epsilon_r$ such that $\alpha$, $\beta$, and $\gamma$ are integers. Object poses can only change through a set of tray tilting actions $A$.
Tilting directions were chosen to make a sequence composed of $N$ actions.
\[ S = \{ a_1, a_2, ..., a_N \}, a_i \in A \]
where $A$ is the set of tilting actions in the cardinal directions.
We execute a sequence $S$ consisting of $N$ random samples from $A$ with replacement, and track the resulting sequence of object poses. We repeat the same sequence $M$ times to obtain an estimated pose probability distribution after each action $a_i$, 
\begin{align}
f^i(x,y,\theta) = \displaystyle \frac{1}{M} V^i_{x,y,\theta}
\end{align}
where $V^i_{x,y,\theta}$ is the number of object poses that occupy the 3-dimensional interval in space, or voxel, $(x,y,\theta)$ after executing action $a_i$. 

Given the pose probability distributions, we can compute the system entropy $H^i$ following action $a_i$.
\begin{align}
H^i = \displaystyle - \sum_{x \in \mathcal{X}} \sum_{y \in \mathcal{Y}} \sum_{\theta \in \Theta} f^i(x,y,\theta) \log_2 f^i(x,y,\theta)
\end{align} 
$H$ can be interpreted as the number of additional information bits required to specify the object pose. 
If the pose distribution is uniform prior to the first tilt, then the entropy would be close to the logarithm of the number of voxels, $H^0=\log_2(\alpha \times \beta \times \gamma)$. 
Ideally, the sequence converges to a fully determined pose, and the entropy drops to zero, $H^N = 0$. In terms of object pose uncertainty, high entropy corresponds to more uncertainty while low entropy corresponds to low uncertainty.

The main experimental challenge is the number of experiments required to reliably estimate the probability distribution of the poses.  Lane~\cite{rice_rule} suggests that the number of trials $M$ should satisfy
\begin{align}
\alpha \times \beta \times \gamma = 2 M^{1/3} \label{rice_rule}
\end{align}
where $\alpha \times \beta \times \gamma$ is the number of voxels. 
The implication is that a large number of trials is required for even a very modest number of voxels. 
For our physical experiments (Section \ref{sec:rob}) we selected a $3 \times 3 \times 3$ grid, which requires $M=2,460$ trials for a high-quality estimate of the probability distribution of the object poses.
We used an action sequence consisting of $N = 50$ tilts, resulting in a total of $M \times N = 123,000$ tilts.

Unfortunately, the object and tray wear down after hundreds of tilts, changing the frictional properties of the system. We therefore settle with $M=500$ trials and a total of $25,000$ tilts. As discussed in Section \ref{sec:disc}, the number of occupied voxels significantly decreases after the first couple of actions. In effect, we have a much smaller number of occupied voxels, which leads us to believe that the smaller number of trials are sufficient for our experiments. 
To conduct experiments on a large scale without the real world challenges such as wear and tear, we look towards simulation.

For analyzing simulation data, we selected a $4 \times 4 \times 4$ grid, which would require $M \approx 32,000$ trials for high quality estimates of the probability distribution of the poses, according to Equation \ref{rice_rule}. 
Our three simulation experiments in Sections \ref{sec:expA},\ref{sec:expB}, and \ref{sec:expC} tested a total of $78$ sequences ($M=10,000$ trials per entropy trend for the first two experiments and $M=1,000$ trials for the third). 
This results in almost $78 \times 32,000 \approx 2,500,000$ trials in total if we were to occupy all $4 \times 4 \times 4 = 64$ voxels across the tested action sequences.
Instead, we conducted $600,000$ trials with sequences consisting of $50$ actions resulting in $30,000,000$ tilts in simulation data.

Note that the effect of our choice of voxel size is reduced by focusing on the change in entropy, rather than the absolute entropy \cite{Chirikjian2009}. 

\begin{figure}[t!]
	\centering
    \includegraphics[width=0.7\linewidth]{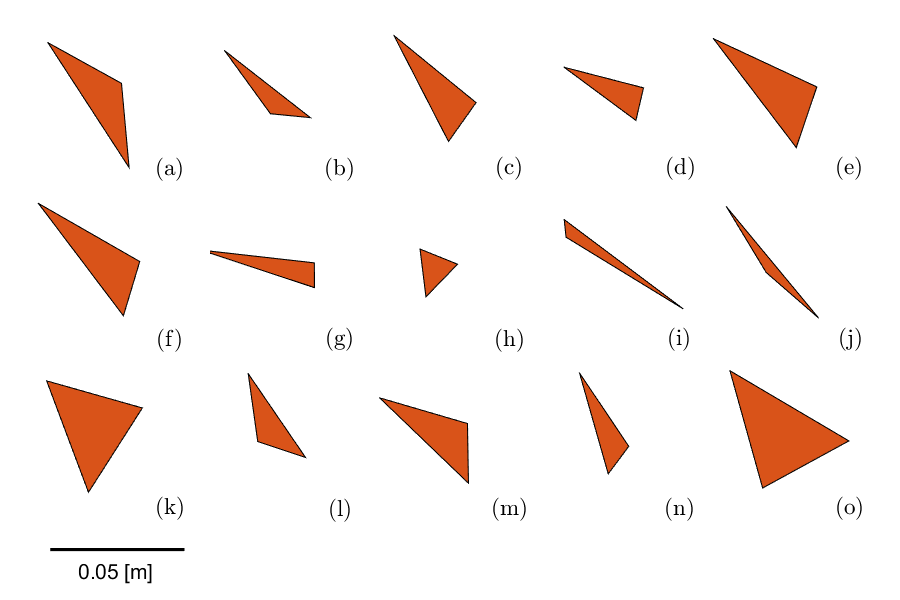}
    \caption{Randomly generated object shapes used in simulation experiments. Density of objects set to be the same as that of the simulated L-shape allen key.}
    \label{fig:obj_shapes}
\end{figure}

\section{Simulation Experiments}
\label{sec:sim}
Executing the experiment first in simulation enables us to generate the necessary number of trials required to estimate the object pose distribution with a sufficient pose resolution, across different action sequences, object shapes, and friction noise levels. 
For a realistic simulation we used the multibody contact friction model library in MATLAB Simscape. 
The tray used in the tray-tilting experiments was modeled as a box with no lid. 
The actions were 30-degree tray tilts in any of the eight cardinal directions. 
We started with an L-shaped object, mimicking the allen key used by Erdmann and Mason~\cite{M_E}. In the rest of the paper, we will use the terms actions and tilts interchangeably.

We simulate the contact model as a linear spring damped normal force with parameters selected to match experimentally observed metal-on-metal interactions. 
In Section \ref{sec:expB}, we used a few other polygonal shapes, as shown in Figure \ref{fig:obj_shapes}. 
All object interactions were modeled similarly, even for varied object shapes.
The friction model is stick-slip with a velocity threshold~\cite{simscape}. 
To simulate noise during sliding for the tray friction noise in Section \ref{sec:expC}, the coeffecient of friction is varied spatially with an amplitude that we can vary to explore the effect of different friction noise levels.

The initial object pose in each trial was sampled from a uniform distribution in the objects configuration space (CSpace). 
Samples where the object was in collision with the wall were rejected.
To compute entropy throughout the sequence of actions as described in Section \ref{sec:entropy}, we discretized the CSpace $(x,y,\theta)$ into $4 \times 4 \times 4 = 64$ $(x,y,\theta)$ voxels.


\begin{figure}[t!]
    \centering
	\includegraphics[width=0.6\linewidth]{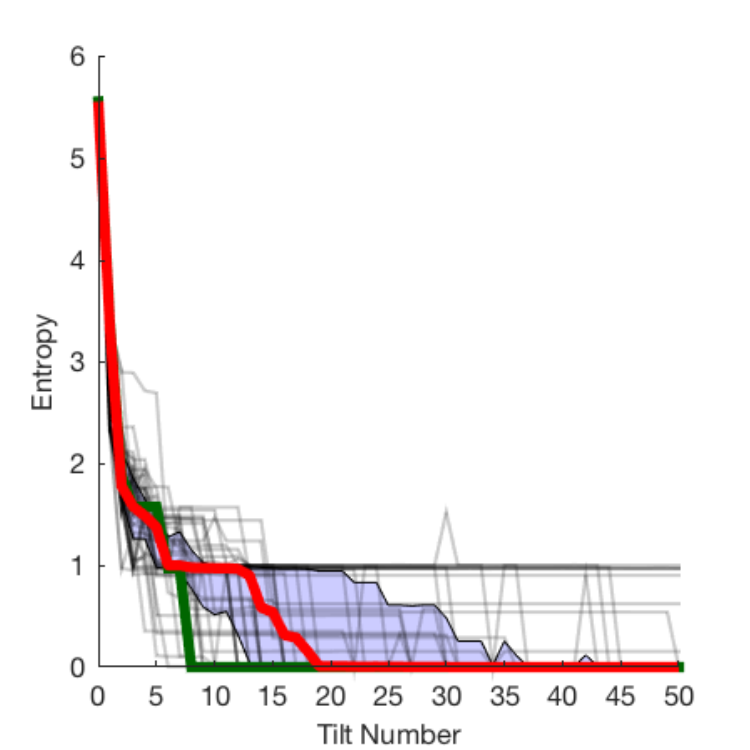}
    \caption{Kruskal effect for the allen key: $M = 10,000$ trials of $N = 50$ actions were repeated across 43 distinct random sequences. The mean (bold red line that converges by the $20^{\text{th}}$ tilt) is bounded by the interquartile range (in blue shaded region). The thin black lines show individual sequences' entropy trends, some of which approach zero by 50 actions. The shortest converging sequence is shown in green reaching zero entropy by the $8^{\text{th}}$ tilt. }
    \label{fig_vary_seq}
\end{figure}

\begin{figure}[ht!]
    \centering
	\includegraphics[width=0.6\linewidth]{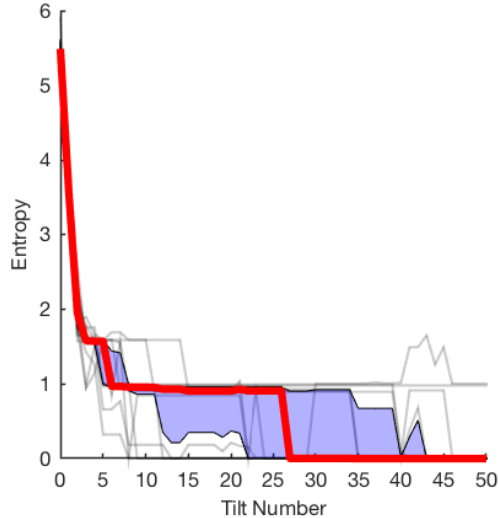}
    \caption{Varying the object shape: $M = 10,000$ trials of $N = 50$ actions were repeated using the same random sequence (green line from Figure \ref{fig_vary_seq}) across 15 various object shapes. The mean (bold red line) is bounded by the interquartile range (in blue shaded region). The thin black lines represent distinct object poses, most of which converge by 50 actions.}
    \label{fig_vary_obj}
\end{figure}

\begin{figure*}[h]
	\centering
    \includegraphics[width=\textwidth]{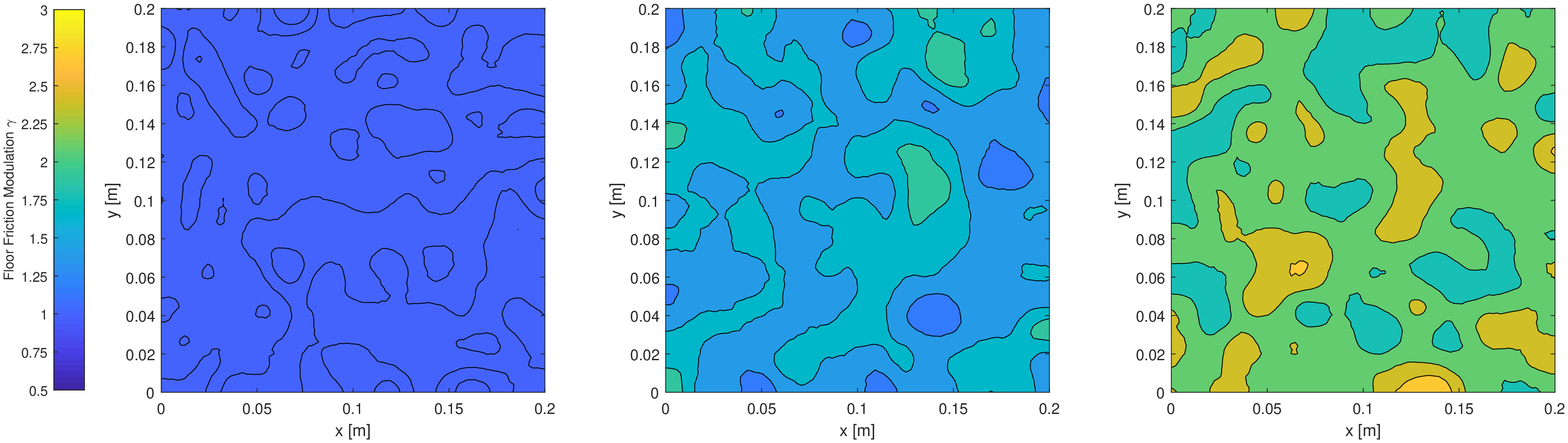}
    \caption{Spatially varying floor friction with low (left), medium (middle), and high (right) variation}
    \label{fig:fric_noise}
\end{figure*}

We conducted three sets of experiments in simulation. Section \ref{sec:expA} tests whether the Kruskal effect could be observed for the L-shaped object.  Section \ref{sec:expB} tested other triangular shapes to confirm that the effect is not specific to L-shaped objects.  Finally, Section \ref{sec:expC} introduced friction noise in our simulation, to observe its effect on convergence rate. Each set of experiments is described below. 

\subsection{Kruskal effect for L-shaped object} 
\label{sec:expA}

The first set of simulation experiments used a single object, the L-shaped model of the allen key. 
We generated 43 distinct random action sequences $S$, each of length $N = 50$. 
Each sequence was repeated $M=10,000$ times, starting from initial poses uniformly sampled from the CSpace as described above. 

Figure~\ref{fig_vary_seq} shows the entropy for each sequence, the mean across all sequences, and the inter-quartile range. In this instance, the Kruskal effect is readily observed.  While the entropy is not monotonically decreasing, there is a clear trend.  Of the 43 sequences tested, 29 converged to zero entropy, with all poses landing in a single voxel. The majority of data ($25\% - 75\%$ or the interquartile range) is within the blue shaded region in Figure \ref{fig_vary_seq}. On average, the entropy converges to a value close to zero by the $20^\text{th}$ tilt. The best randomly-generated sequence converges in eight actions (shown as the green line in Figure \ref{fig_vary_seq}), whereas the Erdmann and Mason plan 
converges in five. While not conclusive, the results suggest that converging plans are common, but optimal plans are rare. This is expected since the actions were randomly chosen instead of being planned.

\subsection{Varying the object shape} \label{sec:expB}
During the second set of experiments, we tested the effect of varying object shape.
We used 15 different triangular object shapes (see Figure \ref{fig:obj_shapes}) and applied the fastest converging sequence we found for the allen key (shown as the green line in Figure \ref{fig_vary_seq}). 
We conducted $M = 10,000$ repetitions for each object, starting at a randomly sampled initial object configuration.

The results are shown in Figure \ref{fig_vary_obj}. Of the 15 objects tested, 10 converged to zero entropy. 
The majority of the data shown by the interquartile range (blue shaded region in Figure \ref{fig_vary_obj}) oriented the test object into a single final determined pose.
On average, the entropy converges to a value close to zero by the $27^\text{th}$ tilt. 

The best sequence generated for the allen key does not perform as well on the other shapes, although it still tends to converge in most cases.
One interpretation is that some objects are harder to orient than others, which is not surprising.  
In the context of pushing, this has already been proven~\cite{Berretty1997}.
It is also likely that we have used a part-specific plan, by generating several part-agnostic plans and then selecting the best for the L-shaped object. We have only restricted to triangular shapes as an initial exploratory experiment, and studying other convex and concave object shapes is left for future work.

\subsection{Varying the friction noise} \label{sec:expC}
The third set of experiments explored the effect of friction noise with the same 30-degree tray tilts and L-shaped object randomly initialized in the CSpace. 
We apply a simple noise model in which we let the  coefficient of friction vary randomly with respect to position within the tray as our real experiments exhibited spatially-varying friction due to wear. 
Figure~\ref{fig:fric_noise} shows the low, medium, and high amplitudes of variation. 
We randomly generated 20 distinct friction maps, which were grouped by their mean friction to generate 13 low-noise maps, 3 medium-noise maps, and 4 high-noise maps. 
We used the allen key, and the best-performing action sequence found for the allen key in the first set of experiments, which is shown as the green line in Figure \ref{fig_vary_seq}. 
The chosen action sequence that converges by eight tilts allows for observable medium and high friction noise convergence behavior, as low friction noise should converge close to eight tilts. 
We performed $M=1,000$ repetitions for each friction map, which is sufficient to observe the effect on the probability distribution of pose between maps.

\begin{figure}[t!]
    \centering
	\includegraphics[width=0.6\linewidth]{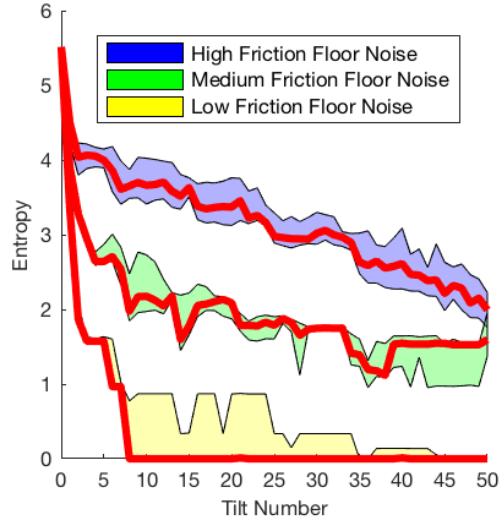}
    \caption{Varying the friction noise: $M = 1,000$ trials of $N = 50$ actions repeated using the same random sequence (green line from Figure \ref{fig_vary_seq}) with 20 distinct floor friction noise amplitudes.  The mean (bold red line) is bounded by the interquartile range for each category of friction floor noise illustrated in Figure~\ref{fig:fric_noise}---low, medium and high.}
    \label{fig_vary_fric}
\end{figure}

The results are shown in Figure \ref{fig_vary_fric}. 
For the low noise maps, 10 of the 13 entropy trends converged to zero entropy. On average, entropy converges to a value close to zero by the $9^\text{th}$ tilt. 

Figure \ref{fig_vary_fric} also shows the results for medium and high noise. 
None of the medium- or high-noise maps converged to zero entropy. 
In both cases, the Kruskal effect is observable in that the general trend of entropy is decreasing, although they tend to not converge to zero entropy within 50 actions. 
These results show that lower friction noise positively affects the probability distribution of object poses towards convergence.  
It is also likely that better-performing sequences exist for higher friction noise levels.
Longer sequences are necessary to draw conclusions as to whether entropy for medium and high friction noise will level off or converge after more than 50 tilts. 

\section{Physical Experiments}
\label{sec:rob}
\begin{figure}[t!]
\centering
\includegraphics[width=0.5\linewidth]{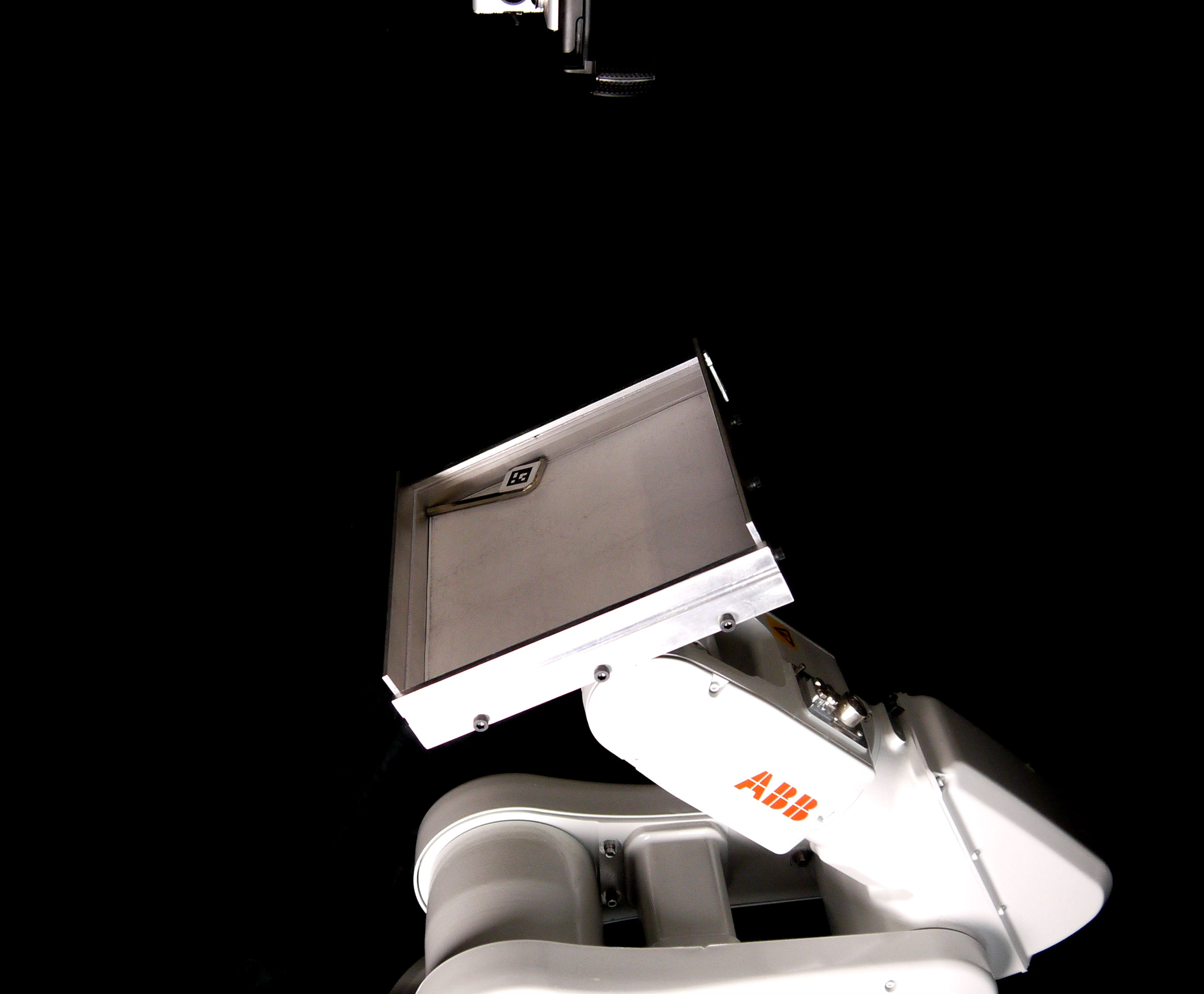}
\caption{Experimental setup.  An industrial robot tilts an allen key, with April Tag attached, in an aluminum tray. The overhead camera records the pose of the allen key after each tilt.}
\label{fig:robot-pic}
\end{figure}

The simulation results suggests that the Kruskal effect can be observed for 2D objects, with significant entropy reduction for a variety of triangular objects and friction noise levels.  
However, physical rigid body interactions can be complex to simulate accurately. 
The goal of the physical experiments is to test the validity of the simulations.
We tested one of the randomly generated sequences consisting of 50 actions.  
We used a 6-DOF ABB IRB 120 robotic arm tilting a 200 mm square aluminum tray.  
The object is a 77.5$\times$27.5 mm allen key with an April Tag \cite{apriltag} to track the object with an overhead camera, as pictured in Figure \ref{fig:robot-pic}. 
The tilting actions were 30-degree tilts in each of the eight cardinal directions.  We ran M=500 trials which is less than the number of trials suggested by Eq. ~\ref{rice_rule}, but due to wearing down of the tray from metal-metal interactions we restrict ourselves to less trials and thus lower resolution ($3 \times 3 \times 3$ grid). This inevitably leads to a less accurate estimate of the entropy, but we still expect to see the downward trend if the Kruskal effect is observed.

It is important that each trial be independent of the preceding trial, and that the initial poses approximate a uniform sampling of the CSpace. 
To that end, the robot shook the tray vigorously prior to the start of every sequence. 
The success of that approach is easily assessed by checking the initial entropy $H^0$. 
A uniform distribution over 27 voxels would yield an entropy of about $4.75$. 
However, finite sampling from a uniform distribution is not likely to yield a uniform distribution. 
Numerical experiments for a dataset of 500 samples drawn into 27 bins suggested an expected initial entropy of approximately $4.72$. 
The measured entropy of our initial distribution is around $3.9$, for a difference of just under one bit. 
We attribute the difference to the fact that some of our CSpace volume $\mathcal{X} \times \mathcal{Y} \times \Theta$ is infeasible due to collisions with walls, and to small limitations in our vigorous shaking motion.

\begin{figure}[t!]
    \centering
	\includegraphics[width=0.6\linewidth]{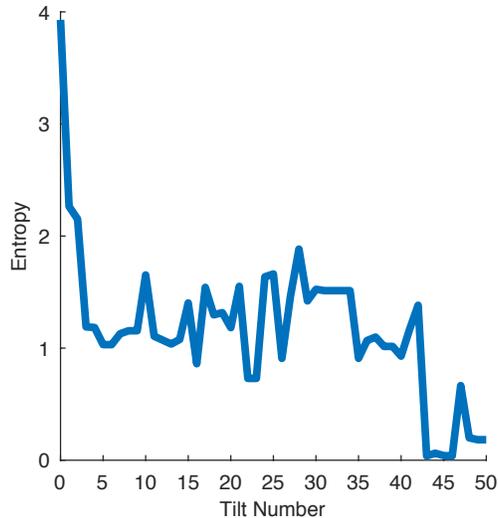}
    \caption{Robot Entropy Data: $M = 500$ trials of $N = 50$ actions repeated on the robot using the same sequence that was used in simulation experiments \ref{sec:expB} and \ref{sec:expC}.}
    \label{fig6}
\end{figure}

Entropy is calculated in the same way as Section \ref{sec:sim}, discretizing the tray volume into $3 \times 3 \times 3 = 27$ $(x,y,\theta)$ voxels. 
Corresponding results are shown in Figure ~\ref{fig6}.
The entropy line is quite noisy which makes it difficult to draw confident conclusions, but the general trend is downwards and indicative of the Kruskal effect. In future work, real world issues like wear and tear should be addressed to obtain more trials and finer resolution for more concrete inferences.

\section{Discussion} 
\label{sec:disc}
In this section, we will discuss the results presented in Sections \ref{sec:sim} and \ref{sec:rob}, draw conclusions and discuss insights for future exploration.

From the planner proposed by Erdmann and Mason~\cite{M_E}, we know that planned actions can orient an allen key to a final determined pose. 
Although their proposed sequence efficiently oriented the object, we wanted to explore how random sequences would perform at the same task. 
Towards this end, Section \ref{sec:expA} tests various random sequences on the same test object. 
Almost all sufficiently long sequences significantly reduce the entropy, and most sequences result in zero entropy. We show that the Kruskal effect applies for any random sequence to mostly or completely reduce object pose uncertainty. 

Given an object, it would be possible to produce an object-specific plan by searching random sequences and selecting the best. 
However, we consider action sequences that are not object-specific which is beneficial when introducing new objects. 
We show this in Section \ref{sec:expB}, where we selected the best allen key sequence, and repeated it for other triangular shapes. 
Figure \ref{fig_vary_obj} shows that the sequence reduced entropy to a few poses within 30 tilts. 
On average, the sequence succeeds at decreasing entropy for all tested objects, perhaps because the objects are all somewhat similar to the L-shaped object. 
Even a part-specific sequence serves as part-agnostic sequence, although a less efficient one. Testing shapes with more edges, especially with a rectangular tray, could affect the amount of uncertainty in object pose. 
A possible extension of this work is to identify such objects and environments. 

In Section \ref{sec:expC}, we explore the significance of non-deterministic actions, by introducing a noise model. 
While the entropy did generally decrease over the tilting sequence regardless of noise, the higher the noise, the slower the object poses seemed to converge. 
The Kruskal effect can be observed in less than ideal conditions such as high noise, but lower friction noises are more efficient at lowering pose uncertainty.
Future work might extend the sequences to see if the entropy levels off at some value depending on the friction noise level or determine whether different sequences perform better at different noise levels.

In Section \ref{sec:rob}, we show that our theory can be applied to the real world. Even with the noise arising from variations in setup and execution, the object poses still converge to a relatively low entropy. In the future we are interested in further exploring the limits of tray tilting actions reducing object pose uncertainty in the real world and the effects of wear on physical systems through exploitation of material interactions. 

Simulation provided large amounts of data and easily varied parameters to confirm the decrease in entropy provided by randomized action sequences. 
The largest entropy decrease among simulation and robot experiments was after the first move. 
At first, random initialization causes the object to be anywhere in the tray and subsequently, only along the edges of the tray after the first tilt.
Testing across different triangular object shapes demonstrated some of the generality in shapes that the system can tolerate. Simulation using different noise parameters, showed that entropy reduction works under stochastic conditions. 

In some of our sequences, the object pose did not converge completely to 0 after 50 iterations. We think this is because for some objects, a certain mini-sequence of actions must be executed consecutively for distinct poses to converge to one pose. When randomly selecting actions, it may sometimes require a very long sequence for this mini-sequence to appear. 
Additionally, for our experiments, small increases in entropy occur due to small changes between similar poses that map to distinct voxels. Later, these poses will converge again but may take some time to find the rights actions to realign.
\begin{figure}[t]
    \centering
    \includegraphics[width=0.7\linewidth]{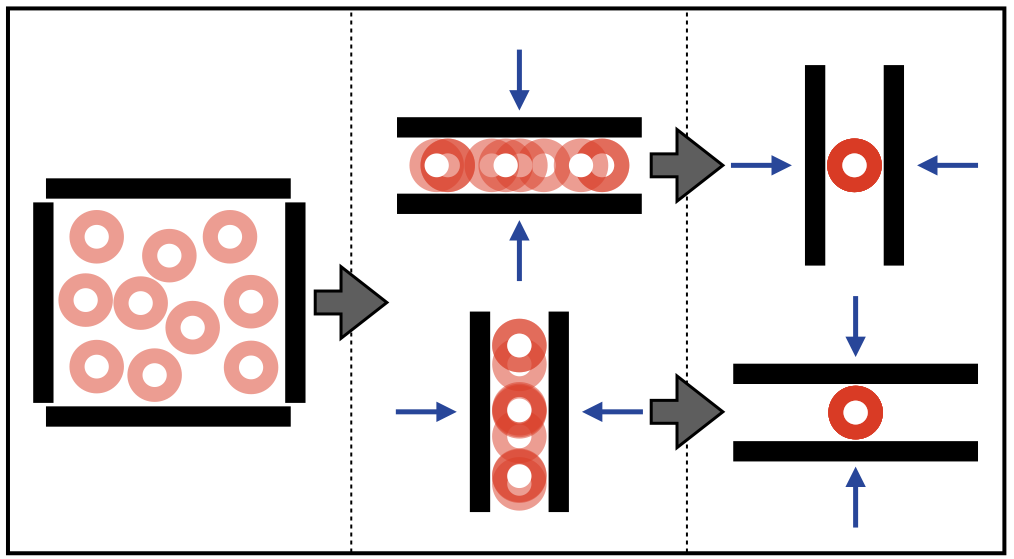}
    \caption{The effect of orthogonal actions on object pose. Long blocks represent a manipulator's two fingers with which we can execute horizontal and vertical squeeze grasps. Translucent disks indicate possible initial poses, and opaque disks indicate the resulting final poses after executing the action. Starting from random initial poses of the disk, the pose uncertainty in (x,y) goes to 0 if the squeeze grasps are orthogonal.}
    \label{fig2}
\end{figure}
Informally, it is possible to make a few observations about the tray tilting process.  The main order-producing phenomenon is when we drive the object pose to the boundary of the CSpace, i.e. a contact between object and tray wall.  Ideally, this is a projection of the feasible poses to the boundary, and reduces the dimension of the feasible CSpace.  For example, if each dimension of $SE(2)$ is quantized into $N$ bins, then at the beginning the pose is spread across $N^3$ bins, and after one action it has been projected to a surface spanned by $N^2$ bins.

In the simplest case, shown in Figure~\ref{fig2} using squeezing actions of a disk, this projection would be a normal projection onto a line. These actions would be analogous to tilting a tray back and forth. For a second action to combine most effectively with the first, the second line would be orthogonal to the first, and the final disk position would then be uniquely determined.

In general, the CSpace surfaces that correspond to kinematic constraints cannot be modeled as linear, nor are the projections linear, but still the toy example may provide some useful insights.  The more closely two actions can be modeled as orthogonal projections, the better.  

The main disorder-producing phenomenon might be sliding across the tray floor, where minor variations in friction can cause rotation of the object.  The vagaries of sliding friction can also make it impossible to say whether an object will stick or slide along a tray wall.

There are also disorder-amplifying phenomena.  For example, if the part strikes the wall sharply it will rebound, and the small variations in initial pose will be integrated over time to produce large variations. It is this effect we relied upon to randomize the object pose prior to testing a sequence of actions in our physical experiments.

The effectiveness of a sequence depends on how common and how effective the order-producing actions are, how frequently combinations occur, and how effectively they combine, versus the frequency and degree of disorder produced by the other actions.  One goal of future work will be to explore this underlying structure more precisely, as a way of characterizing tasks.

\section{Conclusion}
\label{sec:conclusion}
Examining the traditional approach of sense-plan-act, we observe the effects of an alternative approach of executing random sequences of actions without sensing.
We show that a sufficiently long random sequence of actions can move an object from an unknown initial pose to a determined final pose, regardless of initial pose of the object, varying object shapes, and stochasticity in the environment.
This effect is explored in greater detail through simulation using millions of tilts and observing the entropy trends over action sequences. 
We learned how some parameters affect our system: longer sequences lower object uncertainty, and stochasticity in the environment as well as some variation in triangular object shapes does not disturb the system. 
We also illustrated the same effect on a real robot and saw a decreasing trend in entropy. However, the final entropy is not as low as suggested by simulation results, due to real world challenges and complications such as wear and tear.  

This is a different paradigm than the sense-plan-act approach where the final pose and the action sequence to achieve that pose are planned; exploring this alternative paradigm and its limitations could be fruitful. 
We offer insights into the idea of randomized action sequences instead of planning.
The advantage in our setup is that random tray-tilting actions are not part-specific and reduce system complexity for new objects. 
For example, orienting a kit of parts is a hard planning problem, but compartmentalized trays executing random tilting actions is a part-agnostic way to make progress in solving that problem.

\section{Future Work}
\label{sec:future}
Extensions of this work to various polygons or approximations of non-convex objects and tray-tilting alternatives like pushing would further explore the effects of randomized action sequences. The sustainability of our approach can be tested through longer sequences in simulation and on physical systems, as well as more trials for higher quality estimates of entropy. We are also interested in ways to capture the order of the system in a data-efficient way. A future goal is to move towards a tray with a lid that can offer a 3D exploration of part-agnostic tray-tilting to determine 3D object pose. 

To identify action sequences that are efficient at orienting a given object, we could learn a policy like Christiansen et al. \cite{ChrisNomodel}, but with finer discretized tray regions for more accurate object poses.
Another future direction would be to explore potential applications of the proposed approach in simplifying a pose estimation problem for a manipulation task.




\bibliographystyle{spmpsci}
\bibliography{references}
\end{document}